\documentclass[10pt,twocolumn,letterpaper]{article}

\usepackage{cvpr}              

\usepackage[dvipsnames]{xcolor}

\usepackage{comment}

\definecolor{cvprblue}{rgb}{0.21,0.49,0.74}
\usepackage[pagebackref,breaklinks,colorlinks,citecolor=cvprblue]{hyperref}

\def\paperID{41} 
\def\confName{CVPR}
\def\confYear{2024}

\title{CamViG: Camera Aware Image-to-Video Generation with Multimodal Transformers}

\author{Andrew Marmon \and Grant Schindler \and  José Lezama \and Dan Kondratyuk \and Bryan Seybold \and Irfan Essa\\
Google Research\\
{\tt\small \{amarmon|grantschindler|joselezama\}@google.com}
}

\begin{document}
\maketitle


\begin{abstract}
We extend multimodal transformers to include 3D camera motion as a conditioning signal for the task of video generation. Generative video models are becoming increasingly powerful, thus focusing research efforts on methods of controlling the output of such models. We propose to add virtual 3D camera controls to generative video methods by conditioning generated video on an encoding of three-dimensional camera movement over the course of the generated video. Results demonstrate that we are (1) able to successfully control the camera during video generation, starting from a single frame and a camera signal, and (2) we demonstrate the accuracy of the generated 3D camera paths using traditional computer vision methods.
\end{abstract}    
\section{Introduction}
\label{sec:intro}

Generative image and video models are becoming increasingly powerful.  What distinguishes video from imagery is \emph{motion} -- both scene dynamics and implicit camera movement.  Currently these two types of motion are entangled in the video generation process, and we would like to pull them apart by explicitly controlling 3D camera motion during video generation.

The most common interface to video generation today is text prompting.  While text like "zooming in on" and "camera panning" can lead to reasonable results, we propose to add explicit 3D camera controls through a \emph{non-text} input channel to a multi-modal image-to-video generation system. We demonstrate that by conditioning generated video on a \emph{non-text} representation of 3D camera movement over the course of the generated video, we are able to control 3D camera motion during video generation.

\begin{figure}[t]
  \addtocounter{figure}{-1}
  \centering
  \begin{subfigure}{1.0\linewidth}
    \includegraphics[width=1.0\linewidth]{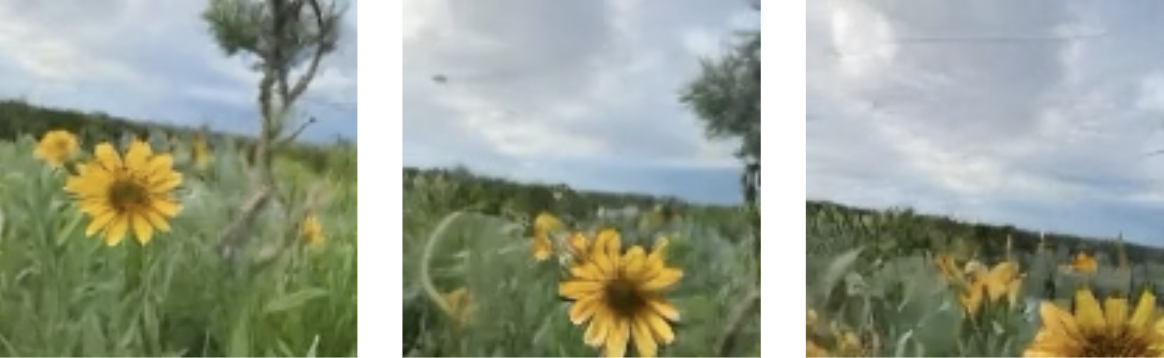}
    \caption{Not camera-conditioned (uncontrolled video generation).}
  \end{subfigure}
  \begin{subfigure}{1.0\linewidth}
    \includegraphics[width=1.0\linewidth]{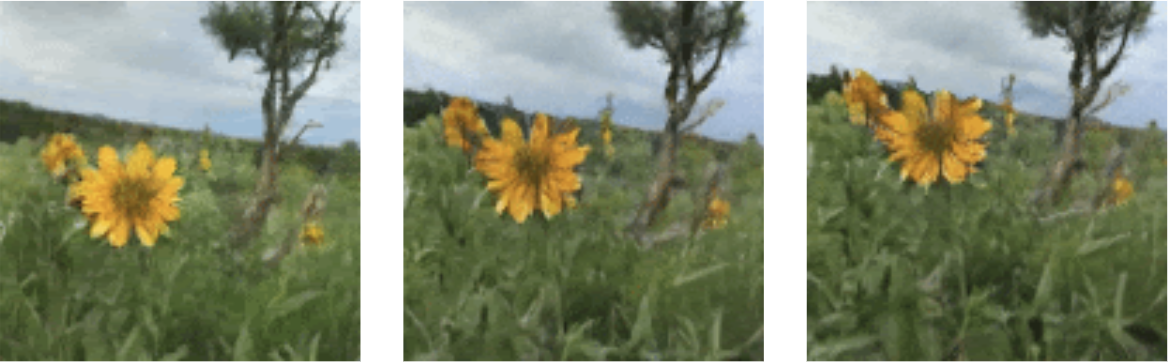}
    \caption{Camera-conditioned (3D translation directly downward).}
  \end{subfigure}
   \label{fig:camera_following}
   \captionof{figure}{Our method generates videos conditioned on a single image and a 3D camera path.  By default, video generation models may move the "implicit camera" viewing the scene in unpredictable ways as in (a). Our method gives explicit control over 3D camera movement, while inducing parallax (despite no explicit 3D or depth representation of the scene) and retaining the inherent inpainting and outpainting abilities of a pre-trained video generation model. In our output (b) above, notice the changing relative positions of flowers against the horizon, and the outpainting at the bottom of the frame as the camera translates downward in 3D.}
   \label{fig:flowers}
\end{figure}   

We are interested in those challenging camera motions, especially 3D camera translations,  which induce parallax and thus require in-painting and out-painting.  Compared to the related field of novel view synthesis, our work has this additional advantage in that in-painting and out-painting come "for free" as part of the generative video pre-training process and require no additional steps to render regions unseen in the single input frame of the video (See Figure \ref{fig:flowers}).

Our primary contribution is an image-to-video generation method that follows camera movement instructions to generate video from a single image of an entire scene.  The method not only (1) shifts the 3D point of view of the scene in a controlled manner, but also (2) allows for scene motion, while (3) automatically performing in-painting and out-painting on all dis-occluded and newly revealed regions in the generated video frames.  We quantitatively evaluate the performance of the camera control method, while providing qualitative comparisons and ablations.

\begin{figure*}
  \centering
  \begin{subfigure}{0.68\linewidth}
    \includegraphics[width=1.0\linewidth]{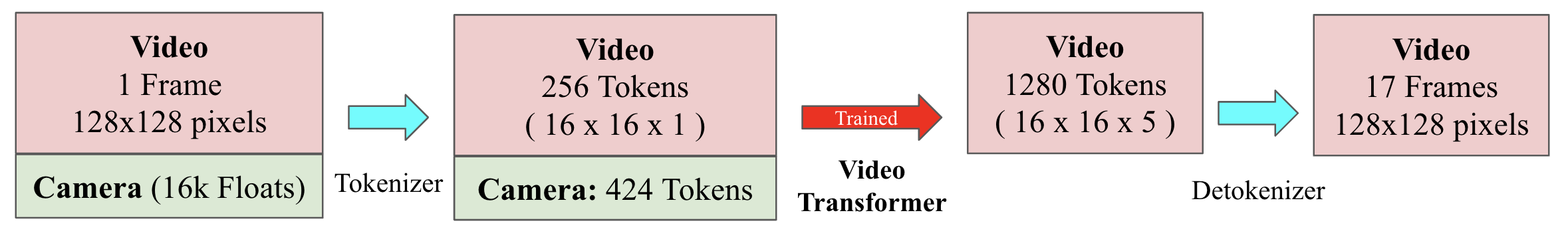}
    \label{fig:short-a}
  \end{subfigure}
  \caption{Model Architecture.  We train a video transformer that takes as input a single frame's worth of visual tokens, plus camera path tokens for the entire length of video.  The output is a set of video tokens that are detokenized into a 17-frame video conditioned on the input camera path.  The tokenizer and detokenizer are frozen, while the video transformer is fully trained to implement our method.}
  \label{fig:short}
\end{figure*}

\section{Related Work}
\label{sec:formatting}

\paragraph{Novel View Synthesis Methods}
Novel view synthesis is a long-standing problem \cite{shum2000} in computer vision, taking as input one or multiple images of a scene, and then outputting an image of the scene from a novel viewpoint. Recently proposed NeRF \cite{mildenhall2020nerf} methods work with as little as a single image, sometimes relying on pre-trained scene priors to reconstruct a scene \cite{yu2020pixelnerf,gu2023nerfdiff}.  SynSin \cite{wiles2020synsin} generates images or videos of a camera trajectory from a single input image, but notably use explicit geometric representations, including learned depth maps and point cloud features, during intermediate steps. DynIBaR\cite{Li_2023_CVPR} tackles dynamic scenes, taking an entire video as input and using neural rendering to synthesize novel views.  Recent methods have also explored a combination of a pre-trained image latent diffusion models with sparse input views \cite{burgess2023viewneti,yoo2023dreamsparse}.

\paragraph{Video-Continuation Methods}
Video-continuation methods such as \cite{52431} are generally not conditioned on camera pose and instead generate the most likely continuation of a video from a single frame or a sequence of short frames. It has been observed that many video-continuation methods default to forward movement into the scene, such as when driving a car \cite{52431}, while our method is explicitly designed to allow movement in all directions even when starting from the same initial video frame.

\paragraph{Object-Centric Methods}
Recently, relative camera pose has been used as a conditioning signal for diffusion-based image generation methods.
Zero-1-to-3 \cite{liu2023zero1to3} fine-tunes a pre-trained Stable Diffusion \cite{Rombach_2022_CVPR} model on image-pose pairs in order to generate pose-conditioned versions of an input image, which they further use for 3D reconstruction.
While their approach is currently restricted to single objects, our method focuses on movement through unrestricted and complex scenes.

\paragraph{Video Generation Models}
A large number of video generation models have recently emerged, including both diffusion-based models \cite{ho2022video} and token-based models \cite{villegas2023phenaki, 52431}. Recent work has extended video diffusion models \cite{wang2023motionctrl} to generate video conditioned on camera pose (as well as other scene motion). In contrast, we use the token-based approach of \cite{kondratyuk2024videopoet}, but build on top of it by introducing the camera path as a new modality.

\section{Approach}
We formulate the problem as follows: Given an input image \emph{I} and a 3D camera path \emph{C}, we generate a set of video frames \(F_1..._n\) such that the first frame of the video \(F_1\) is the input image \emph{I}, and all subsequent frames \(F_2..._n\) follow the camera path specified by \emph{C}.

We use a token-based video transformer model \cite{52431,kondratyuk2024videopoet} as the basis of our camera-conditioned video generation approach such that both the video and the camera path are represented as discrete tokens, as detailed below.  We train our video-transformer on the camera movement task starting from a model that we first pre-train on the video continuation task.

\subsection{Neural Radiance Fields for Data Generation}
We generate ground truth training videos with associated ground truth camera path tokens using a number of house-scale NeRF scenes.  These include apartments, houses, and yards.  For each NeRF scene, we sample 10,000 initial camera poses \emph{(R,T)}.  For each of these initial camera poses, we then generate an additional \emph{n-1} nearby cameras poses along an evenly spaced continuous path.  Rendering these images gives us 10,000 short video clips per scene with known camera paths.

We use synthetic data, and NeRF scenes in particular, for two reasons.  First, rendered NeRF video clips contain true-to-life global illumination and view-dependent lighting effects, as well as finely detailed geometry un-equaled by currently available polygonal synthetic data. Thus, the NeRF renders more closely match the data distribution that the video transformer model has been trained on. Second, we \emph{don't} want to learn only those camera movements which are common in real videos -- generating a distribution of training data which moves equally in all directions for a given scene is intended to remove any movement bias based on the initial image of the scene (e.g. forward motion if the initial image is from a car on a road).

\begin{figure}[t]
  \centering
    \includegraphics[width=1.0\linewidth]{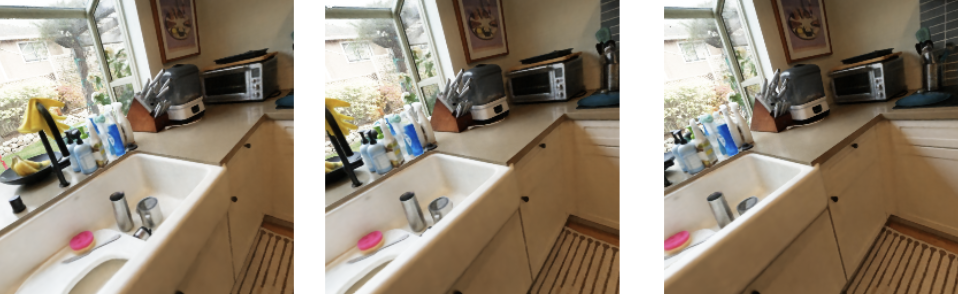}
   \captionof{figure}{NeRF-Based Training Data.  Video sequences with corresponding ground truth camera paths are used to train our model.  We render 10,000 short clips each from large NeRF scenes -- by rendering clips of cameras moving in \emph{all} directions, we remove correlation between scene content and camera motion during training.}
\end{figure}   

\subsection{Video Tokenization and Video Transformer Model}
The video tokenization approach and video transformer model used in this work are based on the approach in \cite{kondratyuk2024videopoet}. Videos are tokenized with a 3D spatial-temporal vector-quantized autoencoder, and processed by an auto-regressive transformer as in \cite{anil2023palm}.  As the focus of this work is specifically on camera control of video generation, we make no changes to the video tokenization step.

\subsection{Representation of Tokenized Camera Paths}
Existing work on multimodal transformers focuses on text, visual, and audio data, with extensive literature surrounding the tokenization of each mode.  Lacking any clear precedent for camera path tokenization, we hypothesized that we could re-use existing neural audio algorithms \cite{zeghidour2021soundstream} to convert camera path data, represented as a 1D array of floating point numbers, into a small number of tokens appropriate for use with our transformer architecture.

The transformer model in \cite{kondratyuk2024videopoet} is a multimodal model that combines video, audio, and text tokens in the same token sequence with special tokens denoting the division between each modality.  We further hypothesized that we could treat our camera tokens exactly like audio tokens, to the extent that the model would use all of its existing audio architecture and special tokens to now represent camera data, both at the level of the camera data tokens and also the special tokens indicating the beginning of the audio signal, which would now mean the beginning of the camera signal.

The largest obstacle to treating camera data as audio was that our pre-trained model was trained on \emph{actual} audio data, so we would be relying on the plasticity of the model to forget about real audio in order to learn that camera path data now occupied the audio channel.

Early experiments confirmed that this approach for both representing and learning camera data as a replacement for audio was successful, and we use it in all experiments detailed below.

\section{Experiments}

\begin{figure}[t]
  \centering
  \includegraphics[width=0.8\linewidth]{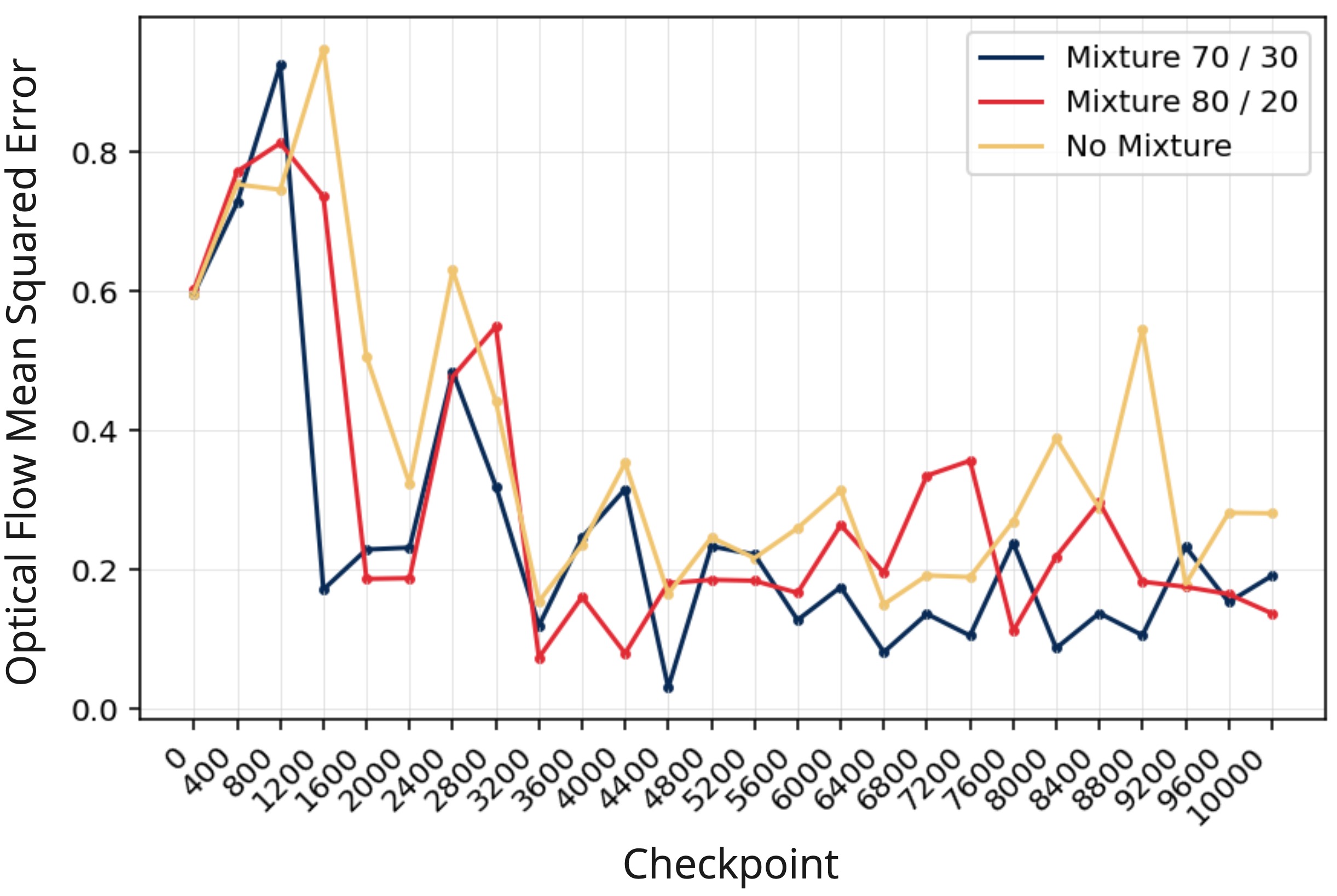}

   \caption{Optical flow MSE (Mean Squared Error) between generated video outputs and ground truth video from the holdout validation set. We show results over the course of training, and for models that use a mixture of 70\% NeRF scenes and 30\% unconstrained video (blue), a mixture of 80\% and 20\% (red), and a model which uses only data from NeRF scenes (yellow). During training, decreasing the difference in optical flow between the generated and ground-truth videos indicates better following of camera movement instructions. We see that as training progresses, overall optical flow MSE decreases. Eventually the model over-fits to the NeRF dataset, resulting in lower quality videos on the unseen validation set which causes a decrease in performance.}
   \label{fig:onecol}
   
\end{figure}

\begin{figure*}[t]
  \centering
    \includegraphics[width=0.85\linewidth]{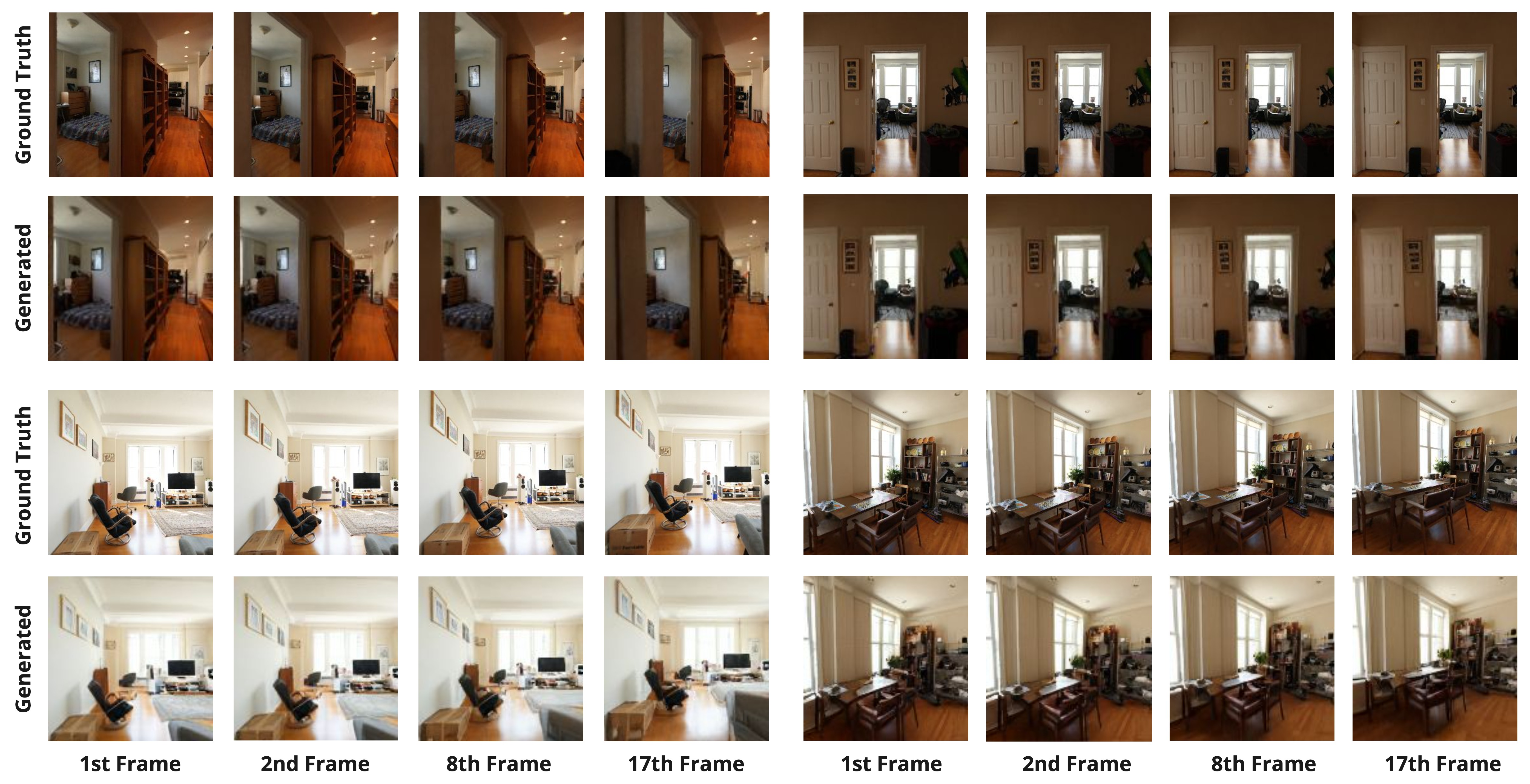}
   \captionof{figure}{A series of four generated videos alongside the ground truth on the holdout validation set. The model is instructed to move the camera left in the top two scenes, and down in the bottom two scenes. We see the model correctly follows the speed and direction instructed while generating convincing outputs. In the first scene (top left), the model correctly induces parallax, flattening the profile of the bookcase and occluding the bed and dresser. We also see correct object generation in the case of finishing the chair legs (bottom right) and the door (top right).}
\end{figure*}

In all experiments, the video transformer model we use is a smaller 1 billion parameter model based on \cite{kondratyuk2024videopoet}. The number of frames \emph{n} per video is 17, and all videos are 128x128 pixel resolution. Each 17-frame video is represented by 1280 video tokens (5x16x16 latent shape) with a video token vocabulary size of \(2^{18}\) as in \cite{kondratyuk2024videopoet}.  Camera paths are represented by 424 tokens.  These 424 tokens are 106 SoundStream \cite{zeghidour2021soundstream} tokens at each of 4 residual vector quantization (RVQ) levels, using a camera token vocabulary of size 4096. The model is trained using a token-based loss as defined in \cite{kondratyuk2024videopoet} for image-to-video generation.

\subsection{Evaluation Metrics}

Error in camera following between a generated video and its associated ground-truth video can be measured by the mean squared error between their respective total optical flows. We measure total optical flow as opposed to per-pixel optical flow as we are generating new pixels at each frame, causing a high per-pixel level loss even when the video is moving in the correct direction. Regardless of per-pixel level variations, however, we expect the total optical flow to be similar across the two videos.

The optical flow parameterized by its x \& y components for any two frames \(F\) is defined as \(O_{x,y}(F_1, F_2)\). The error then between generated frames \(G\) and ground truth frames \(F\) for each video is as follows: 
\[MSE(\sum_{i=1}^{n - 1}O_{x,y}(G_i, G_{i+1}), \sum_{i=1}^{n - 1}O_{x,y}(F_i, F_{i+1}))\]

\subsection{Camera Paths Evaluated}
We define a set of discrete camera paths to evaluate in this work, exploring 3D translation in all cardinal directions (left, right, up, down, forward, backward, and stationary).  The 1D signal \(S\) for each of these paths is defined by the sinusoid  parameterized by the constant \(\lambda\) corresponding with the instructed direction \(d\), with \(\lambda_{d}\in\)\{1,2,3,4,5,6,7\}:

\[S_d = \sin({\pi\lambda_{d}})\]

\subsection{Evaluations}
We evaluate the model's capability to follow a directed camera path by measuring optical flow MSE between generated and ground truth videos on a holdout validation set. For each checkpoint and model variant tested we generate a total of 40 videos to report our results.

As seen in Figure \ref{fig:onecol}, by fine-tuning the model on NeRF scenes with ground-truth camera paths, we are able to guide the model to follow these camera paths during video generation on new frames. We find that the approach that yields the closest camera following with the highest video quality is the model which uses a mixture of 70\% NeRF scene data as well as 30\% large-scale video data which contain no camera controls. The rationale behind this is that while the model learns to leverage the new camera tokens during fine-tuning, it still must maintain its ability to produce realistic video on the generic video dataset.

We observe that, with our current approach, there is a trade-off between learning to reliably move the camera in the intended direction, and maintaining the learned ability of the pre-trained model to generate scene motion. Scenes which  contain motion in the pre-trained model tend to lose motion as the camera becomes more controlled. This is likely due to the NeRF scenes not containing any scene motion, so as the model becomes fine-tuned to this dataset this motion also becomes less likely in general scenes. Using a data mixture during training alleviates this effect, but the resulting models still produce reduced scene motion.
\section{Conclusion}
We demonstrate the ability to treat camera paths as an extra modality in multimodal video transformers, and to generate videos which adhere to precise 3D camera movements defined by input tokenized camera paths. By leveraging a pre-trained backbone and a mixture of training data, this capability extends beyond the interior scenes used for fine-tuning the model, enabling 3D camera control for general image-to-video tasks.
{
    \small
    \bibliographystyle{ieeenat_fullname}
    \bibliography{main}
}


\end{document}